%% file: main.tex
\pdfoutput=1
\documentclass[11pt]{article}

\input{math_commands.tex}

\usepackage[final]{acl}

\usepackage{times}
\usepackage{latexsym}

\usepackage[T1]{fontenc}
\usepackage[utf8]{inputenc}

\usepackage{microtype}

\usepackage{listings}
\usepackage{xcolor}

\usepackage{microtype}
\usepackage{hyperref}
\usepackage{url}
\usepackage{booktabs}
\usepackage{wrapfig,lipsum}

\usepackage{tcolorbox}
\newtcolorbox[list inside=prompt,auto counter,number within=section]{prompt}[1][]{
    fontupper=\ttfamily\footnotesize,
    boxsep=5pt,
    left=0pt,
    right=0pt,
    top=0pt,
    bottom=0pt,
    boxrule=1pt,
    #1,
}

\usepackage{inconsolata}

\usepackage{graphicx}

\title{LLMs Are Not Intelligent Thinkers: Introducing Mathematical Topic Tree Benchmark for Comprehensive Evaluation of LLMs}

\author{{Arash Gholami Davoodi$^1$, Seyed Pouyan Mousavi Davoudi, Pouya Pezeshkpour$^2$} \\
$^1$Carnegie Mellon University, $^2$Megagon Labs\\
\texttt{agholami@andrew.cmu.edu, spouyan.mousavi@gmail.com, pouya@megagon.ai} \\
}

\begin{document}
\maketitle
\begin{abstract}
Large language models (LLMs) demonstrate impressive capabilities in mathematical reasoning. However, despite these achievements, current evaluations are mostly limited to specific mathematical topics, and it remains unclear whether LLMs are genuinely engaging in reasoning. To address these gaps, we present the \textbf{Ma}thematical \textbf{T}opics \textbf{T}ree (MaTT) benchmark, a challenging and structured benchmark that offers 1,958 questions across a wide array of mathematical subjects, each paired with a detailed hierarchical chain of topics. 
Upon assessing different LLMs using the MaTT benchmark, we find that GPT-4 achieved a mere 54\% accuracy in a multiple-choice scenario. Interestingly, even when employing Chain-of-Thought prompting, we observe mostly no notable improvement. Moreover, LLMs accuracy dramatically reduced by up to 24.2 percentage point when the questions were presented without providing choices. Further detailed analysis of the LLMs' performance across a range of topics showed significant discrepancy even for closely related subtopics within the same general mathematical area.
In an effort to pinpoint the reasons behind LLMs performances, we conducted a manual evaluation of the completeness and correctness of the explanations generated by GPT-4 when choices were available. Surprisingly, we find that in only 53.3\% of the instances where the model provided a correct answer, the accompanying explanations were deemed complete and accurate, i.e., the model engaged in genuine  reasoning\footnote{We release our datasets and code at \url{https://github.com/arashgholami/MaTT}}. 
\end{abstract}

%%%%%%%%%%%%%%%%%%%%%%%%
\section{Introduction}
Large Language Models (LLMs) have increasingly demonstrated remarkable capabilities as mathematical reasoners, underscoring their potential in complex problem-solving domains \citep{chowdhery2022palm,touvron2023llama,openai2023gpt-4,team2023gemini}. 
Recent studies have shown that LLMs, when applied to mathematical problems, can exhibit a high degree of reasoning ability, often aligning with or even surpassing human-level performance in certain contexts. 
This proficiency in mathematical reasoning is further enhanced by innovative techniques such as Chain-of-Thought \citep{wei2022chain}, Tree-of-Thought \citep{yao2024tree}, and Self-Verification \citep{weng2022large}, emphasizing on the importance of the procedural steps in solving a mathematical problems.

%%%%%%%%%%%%%%%%%
\begin{figure*}[t!]
    \centering
    \includegraphics[width=0.9\linewidth]{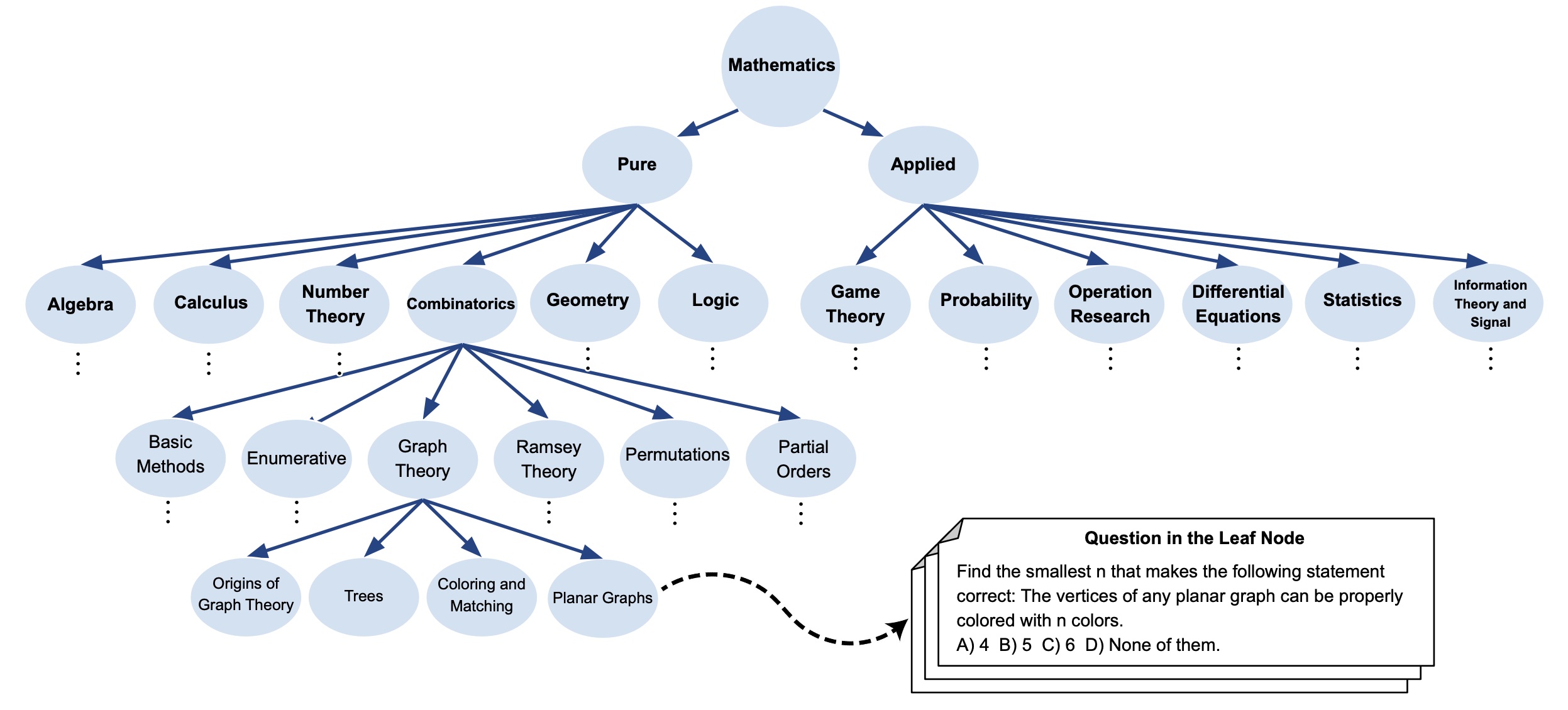}
    \minipostspace
    \caption{Overview of \textbf{Ma}thematical \textbf{T}opics \textbf{T}ree (MaTT) benchmark, a challenging and structured benchmark that presents questions spanning a diverse range of mathematical subjects, each associated with a detailed hierarchical structure of topics.}
    \label{fig:overview}
\postspace
\minipostspace
\end{figure*}
%%%%%%%%%%

Despite these advancements, several critical gaps persist in our understanding of LLMs' mathematical reasoning capabilities.
Firstly, it remains unclear which specific areas of mathematics LLMs excel or falter in, as comprehensive evaluations across diverse mathematical domains are lacking. 
Secondly, distinguishing between instances where LLMs rely on memorization versus genuine reasoning is challenging, raising questions about the depth of their understanding. 
Thirdly, the influence of multiple-choice formats on LLM behavior is not well understood, suggesting that models' performance might be affected by the structure of the questions posed. 
These gaps underscore the necessity for a more robust benchmark that facilitates a holistic evaluation of LLMs, enabling us to dissect their strengths, weaknesses, and the nuances of their problem-solving strategies.

In this paper, we developed the Mathematical Topics Tree (MaTT) benchmark by initially leveraging Wikipedia's ``Lists of mathematics topics''\footnote{\url{https://en.wikipedia.org/wiki/Lists_of_mathematics_topics}} to identify key areas in mathematics, resulting in twelve major topics that span both pure and applied mathematics. This was followed by extracting important reference books for each topic from Wikipedia to build a detailed topical tree. We then further refine the benchmark by using the books' tables of contents to structure a comprehensive tree reflecting the hierarchical organization of mathematical knowledge. Upon completing the topic tree, we extracted questions from the subsections of these books, and gathered them under leaf nodes. Finally, we pair each question with multiple-choice options, enhancing the benchmark's utility for evaluating mathematical understanding. An illustration of MaTT is depicted in Figure \ref{fig:overview}.

After developing MaTT, we evaluate the mathematical reasoning capabilities of various LLMs, including commercial models OpenAI o1-mini, GPT-4 \citep{openai2023gpt-4} and ChatGPT \citep{kocon2023chatgpt} (turbo versions), alongside the open-source LLMs, Llama3.1-70B \citep{touvron2023llama} and Mistral \citep{jiang2023mistral}. Notably, GPT-4 achieved only 54\% accuracy in a multiple-choice format. Furthermore, the use of Chain-of-Thought prompting mostly did not enhance LLMs' performance, underscoring the benchmark's complexity and suggesting that mere step-by-step reasoning might be insufficient.
Also, When questions were presented without multiple-choice options, LLMs accuracy dropped by up to 24.2 points. Our analysis also revealed notable discrepancies in performance across topics, showing inconsistent abilities even within related subtopics of the same mathematical domain.
% when questions were presented without multiple-choice options, we observe a dramatical drop of up to 24.2 percentage point in LLMs accuracy. Additionally, our comprehensive analysis of LLMs' performance across different topics revealed notable discrepancy, highlighting the models' inconsistent ability to address even related subtopics within the same mathematical domain. 

To understand the underlying causes of the LLMs' inadequate performance and their inconsistent results across various topics, we did a detailed evaluation of the explanations provided by GPT-4. Surprisingly, we observe that only in 53.3\% of cases where the models answered correctly, the explanations were also complete, i.e., GPT-4 engaged in genuine reasoning.
These cases were typically associated with simpler or more well-known questions that required only a few straightforward steps to resolve. 
For more complex questions demanding either more number of steps, complicated calculations, or creative/intelligent problem-solving, GPT-4 often failed, resorting to tactics like choice engineering, unsupported theorems, circular reasoning, or memorization instead of true reasoning.

\section{MATT: Mathematical Topics Tree Benchmark}
In recent years, LLMs have shown remarkable abilities in mathematical reasoning. Yet, their prowess is not fully understood due to the narrow focus of current benchmarks, which typically concentrate on specific mathematical areas. This limitation hinders our understanding of the depth and breadth of LLMs' reasoning capabilities. There's a pressing need for more comprehensive mathematical benchmarks that cover a wider array of topics and offer deeper insights into the models' reasoning processes. Such benchmarks would not only challenge the models across a broader mathematical spectrum but also can help with better understanding the nuances of how and where these models apply reasoning.

To address this gap, we create the Mathematical Topics Tree (MaTT) benchmark. 
We start by harnessing the ``Lists of mathematics topics'' available on Wikipedia as a foundational resource. This exploration was crucial for identifying the spectrum of mathematical knowledge we aimed to encompass. Extracting the list of mathematics topics from Wikipedia, we identified twelve principal topics that comprehensively encapsulate the breadth of pure and applied mathematics. Then, for each topic, we extracted one or few key reference books listed on their respective Wikipedia pages. The topics and their corresponding resources are as follows: for pure math we consider Algebra \citep{meyer2023matrix,herstein1991topics,McGee2002logic}, Calculus and Analysis \citep{stewart2012calculus}, Number Theory \citep{niven1991introduction}, Combinatorics \citep{bona2002walk}, Geometry and Topology \citep{coxeter1969introduction,coxeter1967geometry, engelking1989general}, and Logic \citep{mendelson2009introduction}. In applied math we have Game Theory \citep{osborne1994course}, Probability \citep{tijms2012understanding,tijms2017probability}, Operations Research \citep{hillier2015introduction}, Differential Equations \citep{boyce2021elementary}, Statistics \citep{hogg2013introduction}, and Information Theory and Signal Processing \citep{cover1999elements, proakis2007digital}.

Next, we used the tables of contents from selected reference books to structure the MaTT topical tree, mapping the hierarchy of topics and subtopics to create a comprehensive graph of mathematical domains.
% we utilized the tables of content from these selected reference books to enrich and structure the MaTT topical Tree. This approach allowed us to map out the hierarchical organization of topics and subtopics as presented in these books, thereby creating a comprehensive graph that reflects the depth and interconnectivity of mathematical domains. 
The final step in the creation of MaTT involved a detailed extraction of questions from the sections of the reference books, gathering them under the leaf nodes within our topic tree. 
For each question identified, we then crafted multiple-choice options to facilitate an objective assessment framework. 
To generate the options, we selected choices that closely resembled the actual answer, such as those with similar numerical values, those attain by omitting a step from the proof, or those presenting alternative combinations. For instance, if the correct answer was ``A \& B'', we included ``A or B'' as one of the possible choices. We provide an illustration of MaTT in Figure \ref{fig:overview}.

To create the MaTT benchmark, the process involved the following steps: (1) The authors of the paper manually extracted questions and answers from reference books. (2) Another author, independent from the one who extracted the question, reviewed and rechecked the correctness of the question and its answer, revising it if necessary to ensure the quality of the benchmark. During the review process, we observe that around 95\% of the questions showed agreement between annotators, requiring no further revisions. This exhaustive process collectively took the authors more than 300 hours to complete, underscoring the dedication involved in the MaTT dataset creation. 

The statistical overview of the MaTT benchmark is detailed in Table \ref{tab:stat}. The benchmark comprises 1,958 examples, meticulously curated across 12 distinct mathematical topics that span the breadth of pure and applied mathematics. In assembling these questions, we aimed to ensure a broad yet consistent spectrum of difficulty across all topics. To ensure difficulty consistency across various mathematical topics, we anchored our selection process to standardized undergraduate/graduate-level curricula commonly adopted by leading universities. By referencing syllabi and textbooks from comparable degree programs, we maintained a uniform educational framework. This approach allowed us to align the complexity of questions with the expected proficiency of students at the same academic level, regardless of the specific mathematical domain. Consequently, each topic's difficulty was calibrated to match the overall academic standard, promoting consistency throughout the benchmark.

Moreover, we strategically curated a diverse mix of problem types—ranging from computational exercises to proof-based questions and applied scenarios—across all mathematical topics. By balancing the cognitive demands required to solve these problems, we achieved a uniform difficulty level. This methodological diversity ensured that each topic was evaluated on similar grounds, preventing any particular area from being inherently easier or more challenging due to the nature of its questions. Finally, while extracting questions, we exclude questions that are overly popular/simplistic or had their responses provided in the books immediately following them to mitigate the risk of data contamination.

%%%%%%%%%%%%%%%%%%%%
%%%%%%%%%%%%%%
\begin{table*}
\small
\centering
\begin{tabular}{lrrrrr}
\toprule 
&Topics&\# Nodes &\# Leaf &\# Qs&\# Avg leaf's Qs\\
\midrule
\multirow{6}{*}{\rotatebox[origin=c]{90}{\bf Pure Math}}&Algebra& 69 & 49 & 120& 2.45\\
&Calculus and Analysis& 137 & 115 & 517& 4.50\\
&Number Theory& 37 & 31 & 126& 4.06\\
&Combinatorics& 19 & 15 & 139& 9.27\\
&Geometry and Topology& 93 & 81 & 159& 1.96\\
&Logic& 23 & 18 & 35& 1.94\\
\midrule
\multirow{6}{*}{\rotatebox[origin=c]{90}{\bf Applied Math}}
&Game Theory& 23 & 15 & 35& 2.33\\
&Probability& 113 & 91 & 276& 3.03\\
&Operations Research& 64 & 53 & 104& 1.96\\
&Differential Equations& 70 & 60 & 157& 2.62\\
&Statistics& 56 & 48 & 109& 2.27\\
&Information Theory and Signal Processing & 69 & 50 &181& 3.62\\
\midrule
&All& 772 & 625 & 1958 & 3.13\\
\bottomrule
\end{tabular}
\caption{Data Statistics of MaTT. }
\label{tab:stat}
\postspace
\minipostspace
\end{table*}
%%%%%%%%%%%%%%%%%%%%
%%%%%%%%%%%%%%%%%%%%
\section{Experimental details}
We assessed the performance of commercial LLMs—OpenAI o1-mini, GPT-4 \citep{openai2023gpt-4} and ChatGPT \citep{kocon2023chatgpt} (turbo versions)—alongside the open-source LLMs Llama3.1-70B \citep{touvron2023llama} and Mistral \citep{jiang2023mistral} (Mistral-7B-Instruct-v0.2), using the MaTT benchmark. In our evaluation, we structured the prompts to request that LLMs first generate an explanation and then the final answer. In the multiple-choice setting, we specifically directed the models to select one of the provided options (A, B, C, or D) as their final answer. Additionally, for zero-shot chain-of-thought prompting, we appended ``let's think step by step'' to the prompt. 
Examples of the prompts utilized in our experiments are provided in the Appendix \ref{details_prompts}.

%%%%%%%%%%%%%%%%%%%%
\section{Experiments}
In this section, we begin with an analysis of LLMs' mathematical reasoning capabilities using the MaTT benchmark. Subsequently, we examine the variation in model performance across different sub-topics. We then assess the effect of choice availability by presenting MATT questions to LLMs without multiple-choice options. Lastly, we concentrate on GPT-4's explanations, manually annotating the level of reasoning in each explanation and exploring the strategies employed by GPT-4 to arrive at correct answers.

\subsection{LLMs Performance on MaTT}
We present the accuracy of LLMs on the MATT benchmark across various topics in Table \ref{tab:perf}. The performances of models are notably low (except for o1-mini), with GPT-4 achieving only about 54\% accuracy and Mistral performing close to the random choice selection. A detailed examination reveals that Mistral often refuses to answer, claiming the correct option is not listed, while other models try to select the closest match or reason with the available choices when their answer is missing. 
% frequently declines to answer, asserting that the correct choice is not among the provided options, while other models attempt to select the closest match or engage in some form of reasoning with the available choices when their calculated answer is not listed.

Additionally, there is a significant variance in the accuracy levels of LLMs across different topics, with gap as high as 41.2\%, highlighting a significant level of difference in understanding and reasoning capability of LLMs across various mathematical areas. 
Moreover, despite demonstrating strong performance on existing math benchmarks, o1-mini shows room for improvement in the multiple-choice setting of MaTT.
Furthermore, a breakdown of o1-mini’s performance by topic reveals that it excels in areas requiring straightforward problem-solving steps, such as Algebra, Calculus and Analysis, and Number Theory. However, it has more room for improvement in topics demanding higher levels of creativity, such as Combinatorics, Logic, and Game Theory.
Additionally, for topics like Operations Research, Statistics, Information Theory and Signal Processing, we suspect that o1-mini was trained on a smaller amount of data, which may contribute to its lower performance in these areas. 
Finally, we observe that zero-shot CoT prompting mostly did not enhance model performance, potentially due to the complexity of the questions. Many of question in MaTT, require intricate/numerous steps or necessitate intelligent/creative thinking, which cannot be addressed by merely following a few simple steps (For further discussion on other reasoning/prompting strategies, e.g, Program of Thought \cite{wang2022self}, refer to Appendix \ref{PoT}.). 
This observation raises questions about the assumption that CoT prompting is effective in many reasoning tasks. Many available evaluation benchmarks on reasoning tasks are designed to be solved in a few straightforward steps \citep{srivastava2022beyond}, whereas real-world reasoning often involves many steps and requires creative problem-solving. 
%%%%%%%%%%%%%%
\begin{table*}
\small
\centering
\begin{tabular}{lrrrrrrrrrr}
\toprule
&\multirow{2}{*}{\bf Topics} & \multicolumn{2}{c}{\bf GPT-4}&  \multicolumn{2}{c}{\bf ChatGPT} & \multicolumn{2}{c}{\bf Mistral}& \multicolumn{2}{c}{\bf Llama3.1}& \bf o1-mini\\
\cmidrule(lr){3-4}
\cmidrule(lr){5-6} 
\cmidrule(lr){7-8} 
\cmidrule(lr){9-10} 
\cmidrule(lr){11-11} 
&&w/o CoT&w CoT&w/o CoT&w CoT&w/o CoT&w CoT&w/o CoT&w CoT&w/o CoT\\
\midrule
\multirow{6}{*}{\rotatebox[origin=c]{90}{\bf Pure Math}}&Algebra&71.1&73.6&45.5&52.1&33.9&39.7&65.8&69.2&89.7\\
&Calculus&52.2&50.9&41.6&42.6&19.3&19.3&50.6&52.7&88.3\\
&Number Theory&52.4&50.0&54.0&47.6&22.2&23.8&53.7&53.1&84.9\\
&Combinatorics&52.1&55.6&45.1&40.8&21.8&19.0&51.4&50.6&73.1\\
&Geometry&53.8&53.8&51.9&50.0&26.3&27.5&56.0&61.1&71.9\\
&Logic&62.9&65.7&31.4&34.3&34.3&28.6&55.8&61.7&79.4\\
\midrule
\multirow{6}{*}{\rotatebox[origin=c]{90}{\bf Applied Math}}
&Game Theory&40.0&40.0&31.4&45.7&14.3&20.0&54.2&48.5&48.5\\
&Probability&50.5&46.2&36.5&37.9&20.2&17.6&54.3&54.3&75.0\\
&OR&40.6&45.3&37.7&30.2&22.6&24.5&41.3&50.0&66.3\\
&Differential&53.5&52.2&41.5&43.4&18.9&16.3&53.5&57.9&78.9\\
&Statistics&63.3&59.6&56.9&52.3&28.4&23.9&66.9&72.4&77.9\\
&Info and Signal&59.3&53.3&38.2&38.2&29.1&26.6&50.8&61.3&75.1\\
\midrule
&All&54.0&52.7&42.9&42.7&23.1&22.5&53.5&56.6&79.2\\
\bottomrule
\end{tabular}
\caption{Accuracy of LLMs over MaTT benchmark (for accuracy with confidence interval refer to Appendix \ref{CI}).}
\label{tab:perf}
% \minipostspace
\end{table*}

%%%%%%%%%%%%%%%%%%%%%%%
\subsection{Per-Topic Break Down of LLMs Performance}
\begin{figure*}[t]
    \centering
    \includegraphics[width=\linewidth]{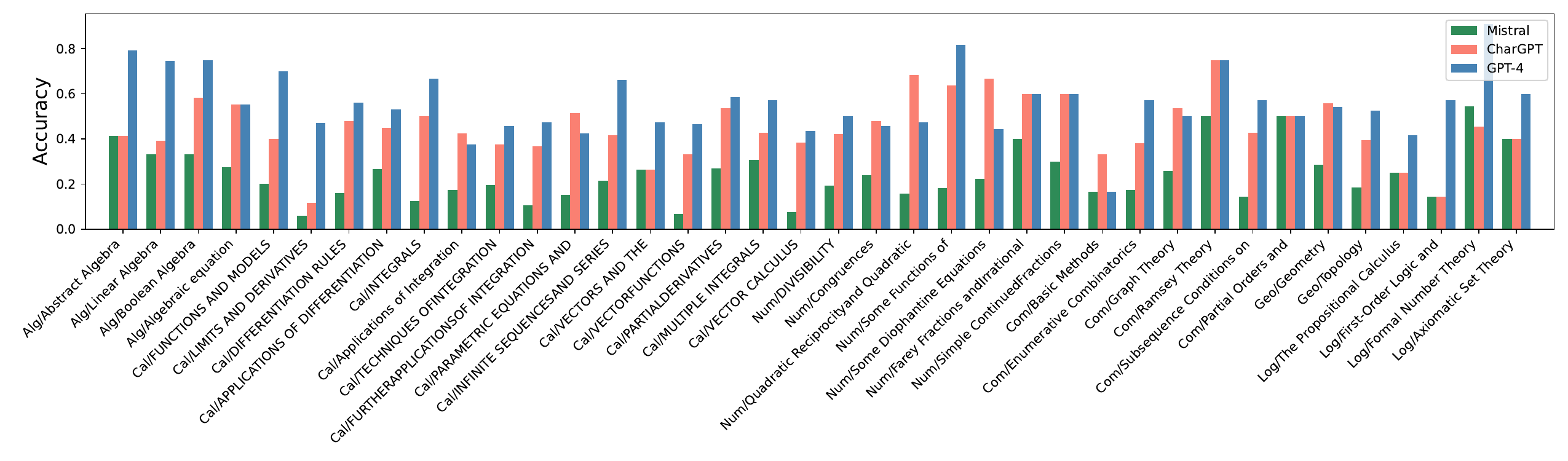}
    \postspace
    \caption{Per-topic breakdown for pure Math.}
    \label{fig:cat-pure}
\minipostspace
\minipostspace
\end{figure*}

\begin{figure*}[t]
    \centering
    \includegraphics[width=\linewidth]{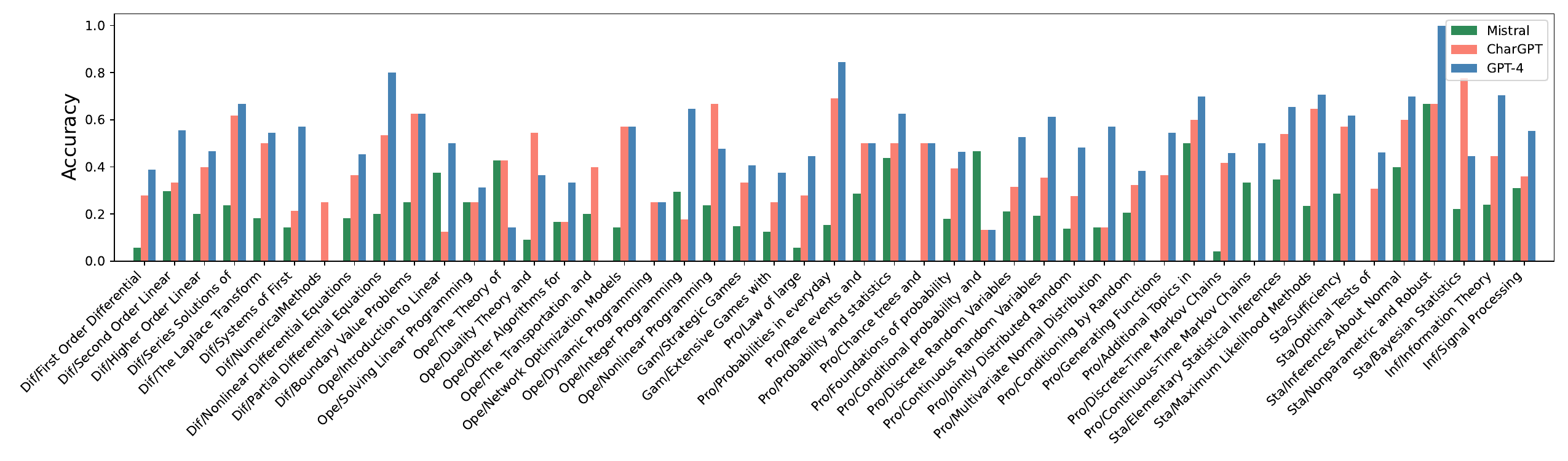}
    \postspace
    \caption{Per-topic breakdown for applied Math.}
    \label{fig:cat-applied}
\minipostspace
\minipostspace
\end{figure*}

As highlighted in the previous section, the exploration of LLMs' capabilities in mathematical reasoning across a diverse array of topics or distinct sub-topics within the same mathematical domain remains significantly unexplored. We detail the LLMs' accuracy on sub-topics within the MATT benchmark in Figures \ref{fig:cat-pure} for pure mathematics and \ref{fig:cat-applied} for applied mathematics, respectively. (We provide the performance breakdown of Llama3.1 and o1-mini in Appendix \ref{perf_breakdown}.)

These figures reveal that the models display varying levels of accuracy even within sub-topics of the same main topic, emphasizing the differences in their understanding and reasoning capabilities even across closely related subjects. Notably, we find that in certain sub-topics, such as application of integration, parametric equations, quadratic reciprocity, diophantine equation, duality theory, non-linear programming, conditional probability, continuous-time Markov chains, and basic statistics, ChatGPT and Mistral outperform GPT-4. This observation further underscores the significance of going beyond the overall performance on high-level topics and instead examining model performance on a more granular level to understand their mathematical reasoning skills comprehensively.

%%%%%%%%%%%%%%%%%
%%%%%%%%%%

%%%%%%%%%%%%%%%%%
%%%%%%%%%%
%%%%%%%%%%%%%%%%%%%%%%%%%%%%%%
\subsection{LLMs Performance without Providing Choices}
To delve deeper into the mathematical reasoning abilities of LLMs, we assessed their performance on the MaTT benchmark without the aid of multiple-choice options. We manually evaluated the models' accuracy on MaTT in the absence of choices and provided the results in Table \ref{tab:wo-choice}. The findings indicate a substantial decrease in performance, with GPT-4, ChatGPT, and Mistral loosing 29.4\%, 56.4\%, and 69.7\% of the accuracy they achieved when choices were available, respectively. This significant decline underscores the models' dependency on choices for deriving answers and highlights their limitations in genuine mathematical reasoning. It also stresses the importance of not solely relying on a single overall score to evaluate LLMs' reasoning capabilities. We provide more detailed analysis on the impact of availability of choices on LLMs prediction in Section \ref{sec:obs}.

%%%%%%%%%%%%%%%%%%%%
\subsection{Reasoning Level of the Explanations}
To understand the reasons behind the poor performance of LLMs without providing choices and their varying accuracy across different topics, we manually examined the completeness and accuracy of LLM-generated explanation for their predictions. Given GPT-4's relatively superior performance compared to other evaluated LLMs, our analysis in this section is specifically focused on the explanations generated by GPT-4. Our objective is to identify the percentage of explanations in correctly predicted instances (when choices are available) for each of the following categories: (1) \emph{complete} reasoning, where the explanation is thorough and logical; (2) \emph{choice/weak} reasoning, where the model uses strategies such as leveraging the given options or offers partial reasoning; and (3) \emph{no/wrong} reasoning, where the explanation is incorrect or missing, and the model reaches a conclusion without justification. Additionally, we calculated the percentage of instances (from all cases where GPT-4 answered correctly with choices) in which GPT-4, with \emph{no choice}, still provided a correct answer and delivered a \emph{complete} explanation.

The results of our manual evaluation of explanations for samples where GPT-4 (when choices are available) predicts the correct answer are detailed in Table \ref{tab:exp}. Remarkably, we found that only 53.3\% of the explanations for correctly answered questions were complete, i.e., GPT-4 engaged in actual reasoning, highlighting a significant inconsistency in GPT-4's actual reasoning abilities. Also, we observe varying levels of explanation completeness across different topics, which do not necessarily correlate with GPT-4's overall performance in those topics.  When comparing samples with complete explanations both with and without choices, we notice a significant gap, underscoring that the presence of choices aids the model in better navigating or recalling the reasoning process.  
Furthermore, we note that GPT-4 genuinely engaged in reasoning primarily for simpler or more well-known questions that could be solved through a few straightforward steps, whereas it struggled with questions requiring more complex steps or creative problem-solving, often resorting to different strategies (we explore these strategies in more detail in Section \ref{sec:obs}). This aligns with the observed limited effectiveness of Chain-of-Thought prompting in enhancing the performance of LLMs. We provide more analysis on explanations in the Appendix \ref{Further_analysis}. 

%%%%%%%%%%%%%%
\begin{table*}[t]
\small
\centering
\begin{tabular}{lrrrr}
\toprule 
&\bf Topics &\bf GPT-4&\bf ChatGPT & \bf Mistral\\
\midrule
\multirow{6}{*}{\rotatebox[origin=c]{90}{\bf Pure Math}}&Algebra& 63.6 \textcolor{red!70!black}{(-7.5) }
&32.5 \textcolor{red!70!black}{(-13.0)}&16.9 \textcolor{red!70!black}{(-17.0)}\\
&Calculus and Analysis&49.7 \textcolor{red!70!black}{(-2.5) }
&23.1 \textcolor{red!70!black}{(-18.5)}&7.3 \textcolor{red!70!black}{(-12.0)}\\
&Number Theory&26.5 \textcolor{red!70!black}{(-25.9)}&19.6 \textcolor{red!70!black}{(-34.4)}&6.3 \textcolor{red!70!black}{(-15.9)}\\
&Combinatorics&43.4 \textcolor{red!70!black}{(-8.7) }&25.4 \textcolor{red!70!black}{(-19.7)}&6.6 \textcolor{red!70!black}{(-15.2)}\\
&Geometry and Topology&40.8 \textcolor{red!70!black}{(-13.0)}&34.9 \textcolor{red!70!black}{(-17.0)}&10.9 \textcolor{red!70!black}{(-15.4)}\\
&Logic&60.7 \textcolor{red!70!black}{(-2.2) }&17.9 \textcolor{red!70!black}{(-13.5)}&14.3 \textcolor{red!70!black}{(-20.0)}\\
\midrule
\multirow{6}{*}{\rotatebox[origin=c]{90}{\bf Applied Math}}
&Game Theory&22.6 \textcolor{red!70!black}{(-17.4)}&22.6 \textcolor{red!70!black}{(-8.8)}&9.7 \textcolor{red!70!black}{(-4.6)}\\
&Probability&32.8 \textcolor{red!70!black}{(-17.7)}&12.3 \textcolor{red!70!black}{(-24.2)}&6.3 \textcolor{red!70!black}{(-13.9)}\\
&Operations Research&15.9 \textcolor{red!70!black}{(-24.7)}&6.9 \textcolor{red!70!black}{(-30.8)}&5.0 \textcolor{red!70!black}{(-17.6)}\\
&Differential Equations&25.0 \textcolor{red!70!black}{(-28.5)}&8.3 \textcolor{red!70!black}{(-33.2)}&4.5 \textcolor{red!70!black}{(-14.4)}\\
&Statistics&38.1 \textcolor{red!70!black}{(-25.2)}&12.3 \textcolor{red!70!black}{(-44.6)}&2.1 \textcolor{red!70!black}{(-26.3)}\\
&Info and Signal&28.3 \textcolor{red!70!black}{(-31.0)}&12.1 \textcolor{red!70!black}{(-26.1)}&5.2 \textcolor{red!70!black}{(-23.9)}\\
\midrule
&All&38.1 \textcolor{red!70!black}{(-15.9)}&18.7 \textcolor{red!70!black}{(-24.2)}&7.0 \textcolor{red!70!black}{(-16.1)}\\
% &Average over Topics&37.3&19.0&7.9\\
\bottomrule
\end{tabular}
\caption{LLMs accuracy in answering questions without providing choices. We demonstrate the decrease in LLMs' performance when choices are not provided, compared to when they are, in \textcolor{red!70!black}{red}.}
\label{tab:wo-choice}
% \minipostspace
\end{table*}
%%%%%%%%%%%%%%%%%%%%

\subsection{Observations from Explanations}
\label{sec:obs}
Besides annotating the reasoning level of explanations (as presented in Table \ref{tab:exp}), we also pinpoint the strategies GPT-4 employs to arrive at correct answers, which do not involve reasoning. We summarise these strategies as follows: 

%%%%%%%%%%%%%%
\paragraph{Choice engineering} refers to the strategy where a model, such as GPT-4, manipulates or exploits the available multiple-choice options to determine an answer, rather than relying on a deep understanding or genuine reasoning process. This can be divided to the following cases:
\begin{itemize} 
    \item{\bf Choices use:} In this case, GPT-4 directly uses the choices and chooses the one matching the question the best. For example, in linear programming questions, despite GPT-4 without choice could not answer any of the optimization problems, when choices were available, using this strategy, GPT-4 achieves a high performance on those questions by simply choosing the minimum or maximum values among the choices. 
    \item{\bf Deducing a plausible answer:} In this strategy, instead of actual reasoning, GPT-4 tries to choose the answer by removing choices that are not plausible answers for the question. For a better understanding, consider the following question: Generate X which has a beta distribution with parameters $\alpha$ and $\beta$. \textbf{GPT4's Answer}: ``Option B incorrectly raises \(U_{1}\) and \(U_{2}\) to the powers of \(\alpha\) and \(\beta\), respectively. This does not correspond to any standard method of generating beta-distributed variables and does not make intuitive sense in the context of the properties of the beta distribution... .''' GPT-4 provides similar arguments for the other options and correctly derive the answer but without any actually reasoning.
    \item{\bf Choice expert:} GPT4 seems to have an understanding of how the choices are usually made. For example consider this question: what are the probabilities of events X and Y?  A)1/3,13/27  B)1/3,1/3  C)1/2,1/2  D)None of them. GPT4 was only able to derive the probability of event X to be 1/3, and without any reasoning claimed that probability of event Y could not be 1/3, and then chose A. Note that we also have the choice ``None of them''.
    \item{\bf Middle ground rule:} We observed that GPT4 tends to choose the middle ground if it cannot find the correct answer. For example: what is the expected duration of the game (which is described in the question and is omitted here)? \textbf{GPT4's answer}:  ``Choices A) 16.519 and C) 22.955 seem more plausible, but without the exact calculation or simulation, it's challenging to pinpoint the exact expected duration. However, B) 19.737 strikes a balance between ... Without the ability to perform a detailed calculation, the most reasonable choice would be: Answer: B) 19.737.''
\end{itemize}

% %%%%%%%%%%%%%%%%%%

\begin{table*}[t]
\small
\centering
\begin{tabular}{lr|rrrr}
\toprule 
&\bf Topics&\bf Complete &\bf Choice/Weak& \bf No/Wrong& \bf No Choice Complete\\
\midrule
\multirow{6}{*}{\rotatebox[origin=c]{90}{\bf Pure Math}}&Algebra&80.5&13.8&5.7&43.7\\
&Calculus and Analysis&79.6&10.4&10.0&66.2\\
&Number Theory&26.9&40.3&32.8&26.9\\
&Combinatorics&33.3&45.3&21.3&30.7\\
&Geometry and Topology&20.0&52.9&27.1&15.3\\
&Logic&72.7&27.3&0.0&54.5\\
\midrule
\multirow{6}{*}{\rotatebox[origin=c]{90}{\bf Applied Math}}
&Game Theory&28.6&35.7&35.7&21.4\\
&Probability&40.0&37.9&22.1&32.1\\
&Operations Research&21.4&28.6&50.0&16.7\\
&Differential Equations&40.0&27.0&32.9&30.6\\
&Statistics&43.5&40.6&15.9&34.8\\
&Info and Signal&68.6&16.2&15.2&42.9\\
\midrule
&All&53.3&27.4&19.3&40.7\\
% &Number of corrects&673&1060\\
\bottomrule
\end{tabular}
\caption{Level of reasoning for explanations in instances where GPT-4's answers were correct, when the choices were available. We report the percentage of explanations with \textbf{complete}, \textbf{choice/weak}, or \textbf{no/wrong} reasoning. We also present the percentage of explanations that exhibited \textbf{complete} reasoning when choices were not provided.} 
\label{tab:exp}
\minipostspace
\minipostspace
\end{table*}
%%%%%%%%%%%%%%%%%%
\postspace
\paragraph{Theorem Use} refers to the strategy where a model, such as GPT-4, utilizes a theorem or property which is the actual and main challenging part of the question. Therefore, by applying it directly, it sidesteps complex derivations and reaches the solution without explaining or proving why that is correct. In order to better grasp the idea, consider the following question and answer from GPT4: What is determinant of Vandermonde Matrix \footnote{Vandermonde Matrix is a $n\times n$ matrix with coefficient ${(x_i)}^j$ at the $i$th row and $j$th column for arbitrary values of $x_i$.}? \textbf{GPT4's Answer}: ``The determinant of a Vandermonde matrix is given by $\prod_{0 \le i < j \le n} (x_j - x_i)$.''

\paragraph{Circular reasoning} is a logical fallacy in which the conclusion of an argument is used as a premise to support itself. This strategy is closely related to hallucination \citep{huang2023survey} which can be very challenging to detect. In these cases, the reasoning starts and ends at the same point, with the argument essentially saying, ``A is true because B is true while B is true as A is true''. For example, let us consider the following question: The Fermat point of a triangle, is a point such that the sum of the three distances from each of the three vertices of the triangle to the point is the smallest possible. What is Fermat point for the triangle having an angle $\ge 120$? \textbf{GPT4's answer}: ``The Fermat point is defined for the purpose of minimizing the total distance from the three vertices of the triangle to this point... When one of the angles of the triangle is greater than or equal to 120 degrees, the Fermat point coincides with the vertex of the triangle that has the angle $\geq 120$ degrees. This is because, in such a case, positioning the point at the obtuse-angled vertex minimizes the total distance from the point to the other vertices of the triangle.'' What GPT4 did was a prime example of circular reasoning and in order to see why, let us dig into it: let's define the statements A=``triangle having an angle $\ge 120$'', B=``a vertex being a Fermat point'' and C=``a vertex minimizing the total distance from the three vertices of the triangle''. B and C are equivalent by definition and $A\rightarrow B$ is almost as hard as $A\rightarrow C$. Here, GPT4 instead of proving $A\rightarrow B$ asserts  $A\rightarrow C$ without any proof, and the concludes $A\rightarrow B$. Not so surprisingly, we observe that GPT-4 frequently uses this strategy in Geometry and Topology, as shown by the high percentage of cases involving choice/weak reasoning in Table \ref{tab:exp}.

\paragraph{Blind memorization} refers to LLMs recalling an answer without following the necessary steps or reasoning to reach it. 
% By blind memorization we mean that LLMs sometimes tend to blindly memorize an answer to a question without learning the necessary steps leading into it or the reasoning behind it. 
Let us further illustrate this with following question: Find the expected value of the distance between two points that are chosen at random inside the unit square, unit circle, and equilateral triangle with sides of unit length. For the square, here is \textbf{GPT4's answer}: ``...The exact solution to this problem involves a double integral that accounts for the distances between all pairs of points within the square. The calculation is somewhat complex but well-documented in mathematical literature. The correct answer, derived from these calculations, is approximately 0.521.'' For the circle, here is \textbf{GPT4's answer}: ``...The exact solution to this problem involves complex calculus and is not trivial to derive without advanced mathematical tools. However, it is known from mathematical studies that the expected distance is approximately $128/(45 \pi)$, which is about 0.905... .'' Finally, for the triangle, \textbf{GPT4's answered} as follows: ``...the exact solution to this problem is not straightforward and requires a detailed calculation that involves the geometry of the equilateral triangle and the properties of distances within it...''. These examples show how blind memorization prevents an LLM from reasoning and answering somewhat similar questions.
% These examples are prime illustrations of how blind memorization will disable an LLM in being able to reason and answer somewhat similar questions. 

%%%%%%%%%%%%%%%%%%%%

%%%%%%%%%%%%%%%%%%%%
\section{Related Work}
As LLMs grow stronger, they exhibit significant mathematical reasoning capabilities on existing benchmarks. However, the scope of current evaluation settings is restricted in terms of the breadth of mathematical areas covered and fails to conclusively determine whether these models genuinely engage in reasoning or rely on alternate strategies.

\paragraph{Mathematical Benchmarks} 
Previous research primarily concentrated on developing benchmarks for math word problems---mathematical problems in the form of written description--which typically require only a few steps to solve, often involving basic arithmetic or elementary algebra \citep{ling2017program, cobbe2021training, patel2021nlp}. Additionally, the work in \citet{mishra2022lila} introduced a comprehensive mathematical reasoning benchmark that encompasses 23 varied tasks across four dimensions: mathematical abilities, language format, language diversity, and external knowledge. Furthermore, \citet{zhang2023multi} presented a multi-modal benchmark with a focus on geometry. The most relevant to our study are the MATH  \citep{hendrycks2021measuring} and Theoremqa \citep{chen2023theoremqa} benchmarks. Despite providing mathematical questions on various topics, they have a much narrower scope compared to our benchmark and did not provide a detailed topical breakdown for each question. Additionally, a recent effort \citep{toshniwal2024openmathinstruct} has begun to generate large-scale synthetic mathematical benchmarks for instruction tuning of LLMs.

\paragraph{LLMs and Math}
In recent years, LLMs have shown notable achievements in mathematical reasoning \citep{srivastava2022beyond,liu2023mathematical}. These accomplishments are supported by methods aimed at enhancing LLMs' performance, predominantly through decomposed reasoning. Such strategies, inspired by human problem-solving processes, include providing step-by-step guidance \citep{wei2022chain, yao2024tree, besta2023graph}, employing verification mechanisms to enhance model consistency and accuracy \citep{weng2022large}, and incorporating complex strategies \citep{qi2023art}.

%%%%%%%%%%%%%%%%%%%%
\section{Conclusion}
In this paper, we provide a comprehensive evaluation on mathematical reasoning of LLMs. 
We create the Mathematical Topics Tree (MaTT) benchmark, a systematically organized set of questions covering a wide range of mathematical subjects linked to a detailed topic hierarchy.
% We create the Mathematical Topics Tree (MaTT) benchmark, a challenging and systematically organized benchmark that presents a series of questions covering an extensive range of mathematical subjects, each linked to a detailed hierarchical structure of topics. 
Exploring LLMs accuracy on MaTT, we observe their struggle with a wide range of mathematical topics, particularly when deprived of multiple-choice options. We also observe the discrepancy in LLMs' performance across various topics and the lack of substantial improvement with Chain-of-Thought prompting.
To investigate the gaps in models performances, we manually analysis their explanations in answering the questions. 
We find that in only 53.3\% of the instances where GPT-4 provided a correct answer, the accompanying explanations were deemed complete. Further, we observe that models perform better on simpler problems but rely on alternative strategies for complex ones. This indicates a fundamental gap in LLMs' ability to engage in deep, creative, and complex mathematical thinking. 
We will release code and data for MaTT. 

%%%%%%%%%%%%%%%%%%%%
\section{Limitations}
This study presents several limitations that should be considered when interpreting the findings. Firstly, our evaluation of mathematical reasoning capabilities was conducted on only five widely adopted LLMs using the MATT benchmark. This limited selection of models may not fully represent the diverse capabilities of LLMs. Including a wider range of models in future assessments could provide a more comprehensive understanding of LLMs' mathematical reasoning across various architectures and training paradigms.

Secondly, our methodology for assessing models' reasoning capabilities heavily relied on analyzing their self-generated explanations. While this approach allows us to gauge how models rationalize their answers, it inherently carries potential biases and inaccuracies. The explanations provided by LLMs might not always accurately reflect the underlying reasoning processes and could sometimes be misleading or incomplete. More objective or diverse methods of evaluation might be necessary to gain a clearer and more accurate picture of how LLMs process and solve mathematical problems.

\bibliography{main}

\appendix
\section{Details of Prompts}
\label{details_prompts}
Example prompts utilized for multiple-choice question answering without and with CoT is  provided in prompts \ref{prompt:wc} and \ref{prompt:wcwc}, respectively. Moreover, the example prompt for answering questions without choices is provided in the prompt \ref{prompt:woc}. 

\begin{prompt}[title={\footnotesize\texttt{Example Prompt with Choices}}, label=prompt:wc]
Choose the answer to the question only from A, B, C, and D choices, and express your reason. \\
Question: Find the smallest n that makes the following statement correct: The vertices of any planar graph can be properly colored with n colors.\\
Choices: A) 4  B) 5  C) 6  D) None of them.\\
The output should be in the following format:\\
Explanation: <explanation>\\
Answer: ----
\end{prompt}

\begin{prompt}[title={\footnotesize\texttt{Example Prompt with Choices and CoT}}, label=prompt:wcwc]
Choose the answer to the question only from A, B, C, and D choices, and express your reason. \\
Question: Find the smallest n that makes the following statement correct: The vertices of any planar graph can be properly colored with n colors.\\
Choices: A) 4  B) 5  C) 6  D) None of them.\\
The output should be in the following format:\\
Explanation: <explanation>\\
Answer: ----\\
Let's think step by step.
\end{prompt}

\begin{prompt}[title={\footnotesize\texttt{Example Prompt without Choices}}, label=prompt:woc]
Answer to the question, and express your reason. \\
Question: Find the smallest n that makes the following statement correct: The vertices of any planar graph can be properly colored with n colors.\\
The output should be in the following format:\\
Explanation: <explanation>\\
Answer: ----
\end{prompt}

\section{Accuracy of LLMs over Matt benchmark with Confidence Interval}\label{CI}
In Table \ref{table:CI}, we calculated a confidence interval (CI) for each topic’s accuracy, assuming benchmark scores are Gaussian distributed, similar to the previous works \cite{dubey2024llama}. 
\begin{eqnarray}
CI= 1.96\times \sqrt{\frac{p(1-p)}{n}}
\end{eqnarray}
where $p$ is the observed benchmark score/accuracy and $n$ is number of samples/questions.

\begin{table*}
\small
\centering
\begin{tabular}{lrrrrrrrrrr}
\toprule
&\multirow{2}{*}{\bf Topics} & \multicolumn{2}{c}{\bf GPT-4}&  \multicolumn{2}{c}{\bf ChatGPT} & \multicolumn{2}{c}{\bf Mistral}& \multicolumn{2}{c}{\bf Llama3.1}& \bf o1-mini\\
\cmidrule(lr){3-4}
\cmidrule(lr){5-6} 
\cmidrule(lr){7-8} 
\cmidrule(lr){9-10} 
\cmidrule(lr){11-11} 
&&w/o CoT&w CoT&w/o CoT&w CoT&w/o CoT&w CoT&w/o CoT&w CoT&w/o CoT\\
\midrule
\multirow{9}{*}{\rotatebox[origin=c]{90}{\bf Pure Math}}&Algebra&\shortstack{71.1\\ {\footnotesize $\pm$8.1}}&\shortstack{73.6\\ {\footnotesize $\pm$7.9}}&\shortstack{45.5\\ {\footnotesize $\pm$8.9}}&\shortstack{52.1\\ {\footnotesize $\pm$8.9}}&\shortstack{33.9\\ {\footnotesize $\pm$8.5}}&\shortstack{39.7\\ {\footnotesize $\pm$8.8}}&\shortstack{65.8\\ {\footnotesize $\pm$8.5}}&\shortstack{69.2\\ {\footnotesize $\pm$8.3}}&\shortstack{89.7\\ {\footnotesize $\pm$5.4}}\\
&Calculus&\shortstack{52.2\\ {\footnotesize $\pm$4.3}}&\shortstack{50.9\\ {\footnotesize $\pm$4.3}}&\shortstack{41.6\\ {\footnotesize $\pm$4.3}}&\shortstack{42.6\\ {\footnotesize $\pm$4.3}}&\shortstack{19.3\\ {\footnotesize $\pm$3.4}}&\shortstack{19.3\\ {\footnotesize $\pm$3.4}}&\shortstack{50.6\\ {\footnotesize $\pm$4.3}}&\shortstack{52.7\\ {\footnotesize $\pm$4.3}}&\shortstack{88.3\\ {\footnotesize $\pm$2.8}}\\
&Number Theory&\shortstack{52.4\\ {\footnotesize $\pm$8.7}}&\shortstack{50.0\\ {\footnotesize $\pm$8.8}}&\shortstack{54.0\\ {\footnotesize $\pm$8.7}}&\shortstack{47.6\\ {\footnotesize $\pm$8.7}}&\shortstack{22.2\\ {\footnotesize $\pm$7.3}}&\shortstack{23.8\\ {\footnotesize $\pm$7.4}}&\shortstack{53.7\\ {\footnotesize $\pm$8.7}}&\shortstack{53.1\\ {\footnotesize $\pm$8.7}}&\shortstack{84.9\\ {\footnotesize $\pm$6.3}}\\
&Combinatorics&\shortstack{52.1\\ {\footnotesize $\pm$8.3}}&\shortstack{55.6\\ {\footnotesize $\pm$8.2}}&\shortstack{45.1\\ {\footnotesize $\pm$8.3}}&\shortstack{40.8\\ {\footnotesize $\pm$8.2}}&\shortstack{21.8\\ {\footnotesize $\pm$6.9}}&\shortstack{19.0\\ {\footnotesize $\pm$6.5}}&\shortstack{51.4\\ {\footnotesize $\pm$8.3}}&\shortstack{50.6\\ {\footnotesize $\pm$8.3}}&\shortstack{73.1\\ {\footnotesize $\pm$7.4}}\\
&Geometry&\shortstack{53.8\\ {\footnotesize $\pm$7.8}}&\shortstack{53.8\\ {\footnotesize $\pm$7.8}}&\shortstack{51.9\\ {\footnotesize $\pm$7.8}}&\shortstack{50.0\\ {\footnotesize $\pm$7.8}}&\shortstack{26.3\\ {\footnotesize $\pm$6.8}}&\shortstack{27.5\\ {\footnotesize $\pm$6.9}}&\shortstack{56.0\\ {\footnotesize $\pm$7.7}}&\shortstack{61.1\\ {\footnotesize $\pm$7.6}}&\shortstack{71.9\\ {\footnotesize $\pm$7.0}}\\
&Logic&\shortstack{62.9\\ {\footnotesize $\pm$16.0}}&\shortstack{65.7\\ {\footnotesize $\pm$15.7}}&\shortstack{31.4\\ {\footnotesize $\pm$15.3}}&\shortstack{34.3\\ {\footnotesize $\pm$15.7}}&\shortstack{34.3\\ {\footnotesize $\pm$15.7}}&\shortstack{28.6\\ {\footnotesize $\pm$15.0}}&\shortstack{55.8\\ {\footnotesize $\pm$16.5}}&\shortstack{61.7\\ {\footnotesize $\pm$16.1}}&\shortstack{79.4\\ {\footnotesize $\pm$13.4}}\\
\midrule
\multirow{9}{*}{\rotatebox[origin=c]{90}{\bf Applied Math}}
&Game Theory&\shortstack{40.0\\ {\footnotesize $\pm$16.2}}&\shortstack{40.0\\ {\footnotesize $\pm$16.2}}&\shortstack{31.4\\ {\footnotesize $\pm$15.3}}&\shortstack{45.7\\ {\footnotesize $\pm$16.5}}&\shortstack{14.3\\ {\footnotesize $\pm$11.6}}&\shortstack{20.0\\ {\footnotesize $\pm$13.2}}&\shortstack{54.2\\ {\footnotesize $\pm$16.5}}&\shortstack{48.5\\ {\footnotesize $\pm$16.6}}&\shortstack{48.5\\ {\footnotesize $\pm$16.6}}\\
&Probability&\shortstack{50.5\\ {\footnotesize $\pm$5.9}}&\shortstack{46.2\\ {\footnotesize $\pm$5.9}}&\shortstack{36.5\\ {\footnotesize $\pm$5.7}}&\shortstack{37.9\\ {\footnotesize $\pm$5.7}}&\shortstack{20.2\\ {\footnotesize $\pm$4.7}}&\shortstack{17.6\\ {\footnotesize $\pm$4.5}}&\shortstack{54.3\\ {\footnotesize $\pm$5.9}}&\shortstack{54.3\\ {\footnotesize $\pm$5.9}}&\shortstack{75.0\\ {\footnotesize $\pm$5.1}}\\
&OR&\shortstack{40.6\\ {\footnotesize $\pm$9.6}}&\shortstack{45.3\\ {\footnotesize $\pm$9.7}}&\shortstack{37.7\\ {\footnotesize $\pm$9.5}}&\shortstack{30.2\\ {\footnotesize $\pm$9.1}}&\shortstack{22.6\\ {\footnotesize $\pm$8.3}}&\shortstack{24.5\\ {\footnotesize $\pm$8.5}}&\shortstack{41.3\\ {\footnotesize $\pm$9.6}}&\shortstack{50.0\\ {\footnotesize $\pm$9.8}}&\shortstack{66.3\\ {\footnotesize $\pm$9.0}}\\
&Differential&\shortstack{53.5\\ {\footnotesize $\pm$7.8}}&\shortstack{52.2\\ {\footnotesize $\pm$7.8}}&\shortstack{41.5\\ {\footnotesize $\pm$7.7}}&\shortstack{43.4\\ {\footnotesize $\pm$7.8}}&\shortstack{18.9\\ {\footnotesize $\pm$6.1}}&\shortstack{16.3\\ {\footnotesize $\pm$5.8}}&\shortstack{53.5\\ {\footnotesize $\pm$7.8}}&\shortstack{57.9\\ {\footnotesize $\pm$7.7}}&\shortstack{78.9\\ {\footnotesize $\pm$6.2}}\\
&Statistics&\shortstack{63.3\\ {\footnotesize $\pm$9.2}}&\shortstack{59.6\\ {\footnotesize $\pm$9.3}}&\shortstack{56.9\\ {\footnotesize $\pm$9.4}}&\shortstack{52.3\\ {\footnotesize $\pm$9.5}}&\shortstack{28.4\\ {\footnotesize $\pm$8.4}}&\shortstack{23.9\\ {\footnotesize $\pm$8.1}}&\shortstack{66.9\\ {\footnotesize $\pm$9.0}}&\shortstack{72.4\\ {\footnotesize $\pm$8.7}}&\shortstack{77.9\\ {\footnotesize $\pm$8.4}}\\
&Info and Signal&\shortstack{59.3\\ {\footnotesize $\pm$7.2}}&\shortstack{53.3\\ {\footnotesize $\pm$7.3}}&\shortstack{38.2\\ {\footnotesize $\pm$6.9}}&\shortstack{38.2\\ {\footnotesize $\pm$6.9}}&\shortstack{29.1\\ {\footnotesize $\pm$6.5}}&\shortstack{26.6\\ {\footnotesize $\pm$6.3}}&\shortstack{50.8\\ {\footnotesize $\pm$7.3}}&\shortstack{61.3\\ {\footnotesize $\pm$7.0}}&\shortstack{75.1\\ {\footnotesize $\pm$6.2}}\\
\midrule
&All&\shortstack{54.0\\ {\footnotesize $\pm$2.4}}&\shortstack{52.7\\ {\footnotesize $\pm$2.4}}&\shortstack{42.9\\ {\footnotesize $\pm$2.4}}&\shortstack{42.7\\ {\footnotesize $\pm$2.4}}&\shortstack{23.1\\ {\footnotesize $\pm$2.0}}&\shortstack{22.5\\ {\footnotesize $\pm$2.0}}&\shortstack{53.5\\ {\footnotesize $\pm$2.4}}&\shortstack{56.6\\ {\footnotesize $\pm$2.4}}&\shortstack{79.2\\ {\footnotesize $\pm$2.1}}\\
\bottomrule
\end{tabular}
\caption{Accuracy of LLMs over the MaTT benchmark with confidence intervals.}
\label{table:CI}
% \minipostspace
\end{table*}

\section{Other Reasoning/Prompting Strategies}
\label{PoT}
Program of Thought (PoT) is a prompting technique designed to enhance numerical reasoning in large language models (LLMs) by integrating code generation into the reasoning process. Unlike Chain-of-Thought (CoT) prompting, where the model performs all reasoning and computation within natural language, PoT allows the model to generate executable code (e.g., Python scripts) as part of its output. This approach delegates complex computations, iterations, and algorithmic tasks to an external interpreter, thereby decoupling intricate calculations from the reasoning steps. By leveraging programming constructs, PoT aims to overcome limitations of LLMs in handling arithmetic operations, solving complex equations, and performing extensive iterative processes that are challenging to express and compute accurately in plain text.

In addition to employing CoT, focusing on GPT-4, we applied PoT prompting to several topics in our MaTT benchmark. However, we observed that for most problems in pure and applied mathematics, breaking down a problem into the PoT format is as challenging as solving the original problem itself. Crafting a correct and efficient program requires a deep understanding of the mathematical concepts and the ability to translate them into algorithmic procedures. This translation often demands the same level of insight and problem-solving skills as directly finding the solution, which poses a significant challenge for LLMs. Consequently, while PoT has the potential to handle computational complexity effectively, its benefits are limited when the model struggles with the initial reasoning required to generate appropriate code.

Moreover, we did not see any significant improvement in the multiple-choice accuracy when using PoT, even in cases where converting the problem into PoT format is not inherently difficult. For example, consider the following question: "A stick is broken into  n pieces at n-1 randomly chosen points (n>2). What are the expected values of the lengths of the shortest and longest segments?" Translating the problem into Python code: 

\begin{lstlisting}[language=Python, caption=Python code to compute the expected lengths of the shortest and longest segments when a stick of length 1 is broken at n-1 random points., basicstyle=\footnotesize\ttfamily,linewidth=\columnwidth, numbers=none,
  xleftmargin=0.05\columnwidth,
  xrightmargin=0.05\columnwidth,]
import numpy as np

n = 5  # Example value for n > 2
num_trials = 100000
shortest_lengths = []
longest_lengths = []

for _ in range(num_trials):
    break_points = np.sort(np.random.uniform(0, 1, n - 1))
    segments = np.diff([0] + break_points.tolist() + [1])
    shortest_lengths.append(np.min(segments))
    longest_lengths.append(np.max(segments))

expected_shortest = np.mean(shortest_lengths)
expected_longest = np.mean(longest_lengths)

print("Expected length of the shortest segment:", expected_shortest)
print("Expected length of the longest segment:", expected_longest)
\end{lstlisting}

We are not able to get the answer for general $n$ with the help of Program of Thought.

\section{Performance Breakdown of Llama3.1 and o1-mini}\label{perf_breakdown} 
\begin{figure*}[t!]
    \centering
    \includegraphics[width=1.0\linewidth]{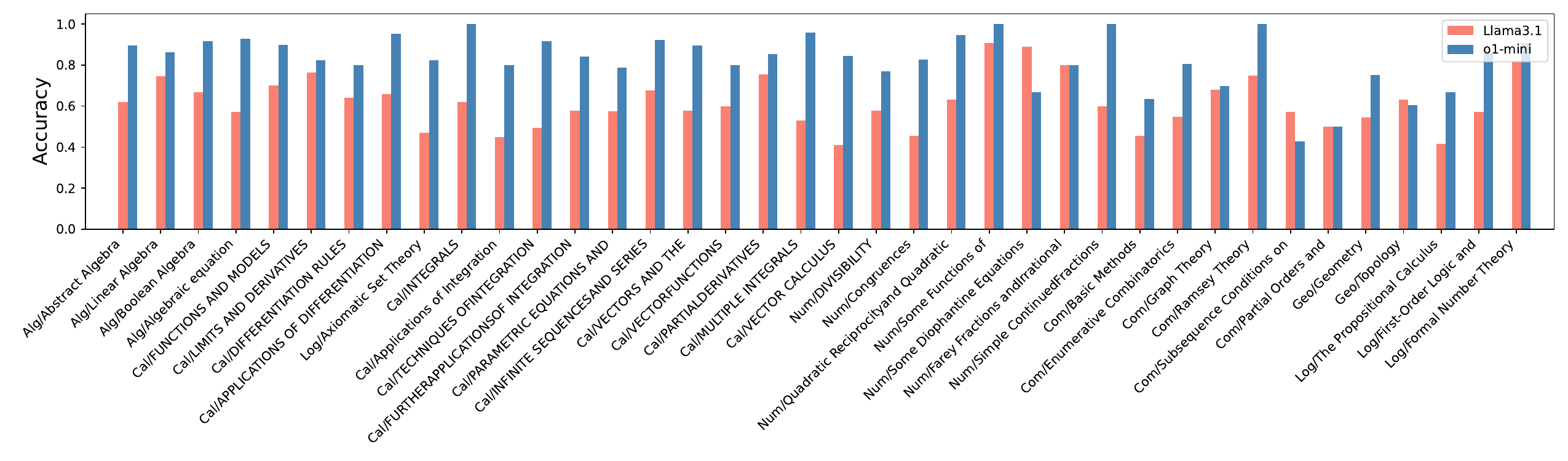}
    \caption{Overview of per topic breakdown for topics under Mathematics/Pure. In this Figure we can observe that in the majority of subtopics (35 out of 38) o1-mini is outperforming Llama3.1, while in the rest of 3 out of 38 subtopics Llama3.1 is outperforming o1-mini.}
    \label{fig:pure_per}
\minipostspace
\minipostspace
\end{figure*}

\begin{figure*}[t!]
    \centering
    \includegraphics[width=1.0\linewidth]{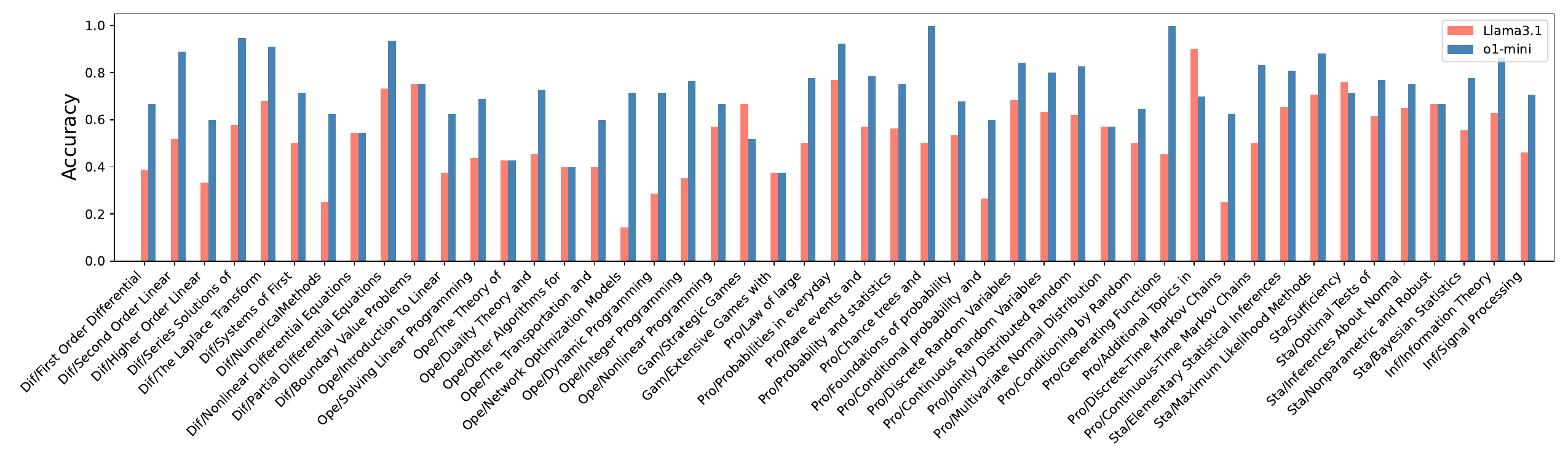}
    \caption{Overview of per topic breakdown for topics under Mathematics/Applied. In this Figure we can observe that in the majority of subtopics (44 out of 47) o1-mini is outperforming Llama3.1, while in the rest of 3 out of 47 subtopics Llama3.1 is outperforming o1-mini.}
    \label{fig:applied_per}
\minipostspace
\minipostspace
\end{figure*}

The performance comparison between o1-mini and Llama3.1 models across topics under both Applied and Pure Mathematics reveals a clear trend in favor of o1-mini. In Pure Mathematics, o1-mini outperforms Llama3.1 in 35 out of 38 subtopics, with Llama3.1 taking the lead in only 3 subtopics (Figure \ref{fig:pure_per}). Similarly, for Applied Mathematics, o1-mini shows dominance in 44 out of 47 subtopics, with Llama3.1 outperforming in just 3 cases (Figure \ref{fig:applied_per}). These results and Table \ref{tab:perf} highlight the consistent superiority of o1-mini across the majority of subtopics in both domains.

%%%%%%%%%%%%%%%%%

%%%%%%%%%%

%%%%%%%%%%%%%%%%
\section{Further Analysis on Explanations}\label{Further_analysis}
To better understand the influence of choices and to distinguish between instances where the model genuinely engages in reasoning, we provided further analysis in GPT-4 generated explanations. We aim to identify the number of samples in which GPT-4 with choices gave a complete explanation, GPT-4 without choices provided a complete explanation, and both scenarios resulted in complete explanations (over all the questions in MaTT). The findings are presented in Table \ref{tab:w/wo-comp}. The result indicates that in most topics, samples that had complete explanations even without the availability of choices also had complete explanations when GPT-4 was provided with choices. Furthermore, in some topics, there is a meaningful difference in the percentage of complete explanations between scenarios with and without choices, emphasizing that the presence of choices can aid models in better engaging with or recalling the reasoning process.

\begin{table*}[t]
\small
\centering
\begin{tabular}{lr|rrrr}
\toprule 
&\bf Topics&\bf both Complete &\bf No Choice Complete& \bf With Choice Complete\\
\midrule
\multirow{6}{*}{\rotatebox[origin=c]{90}{\bf Pure Math}}&Algebra&28.3&36.7&58.3\\
&Calculus and Analysis&30.8&44.7&41.4\\
&Number Theory&4.8&16.7&14.3\\
&Combinatorics&6.5&20.1&18.0\\
&Geometry and Topology&2.5&10.1&10.7\\
&Logic&22.9&42.9&45.8\\
\midrule
\multirow{6}{*}{\rotatebox[origin=c]{90}{\bf Applied Math}}
&Game Theory&8.6&11.4&11.4\\
&Probability&13.4&19.9&20.3\\
&Operations Research&4.8&10.6&8.7\\
&Differential Equations&13.4&22.3&21.7\\
&Statistics&19.3&24.8&27.5\\
&Info and Signal&24.9&27.1&39.8\\
\midrule
&All&18.0&27.4&28.9\\
% &Number of corrects&673&1060\\
\bottomrule
\end{tabular}
\caption{Comparison on the completeness of explanations from GPT-4 when choices were provided versus when no choices were given (this is over all the samples in MaTT).}
\label{tab:w/wo-comp}
\end{table*}
%%%%%%%%%%%%%%%%%%

\end{document}

%% file: math_commands.tex
\usepackage{amsmath,amsfonts,bm}

\def\eqref#1{equation~\ref{#1}}
\def\1{\bm{1}}

\DeclareMathAlphabet{\mathsfit}{\encodingdefault}{\sfdefault}{m}{sl}
\SetMathAlphabet{\mathsfit}{bold}{\encodingdefault}{\sfdefault}{bx}{n}

\usepackage{color}
\usepackage{times}
\usepackage{epsfig}
\usepackage{graphicx}
\usepackage{amsmath}
\usepackage{amsthm}
\usepackage{amssymb}
\usepackage{mathtools}
\usepackage{multirow}
\usepackage{bbm}
\usepackage{dsfont}

\usepackage[table, dvipsnames]{xcolor}

\usepackage[utf8]{inputenc}
\usepackage{booktabs,siunitx}
\usepackage{multirow}
\usepackage{multicol}
\usepackage{amsmath}
\usepackage{subcaption}
\usepackage{caption}
\usepackage{graphicx}
\usepackage{pifont}

\usepackage{tikz}
\usetikzlibrary{shapes,arrows,patterns}
\usetikzlibrary{positioning}
\usetikzlibrary{calc}

\newcommand{\cut}[1]{}

\newcommand{\postspace}{\vskip -3mm}
\newcommand{\minipostspace}{\vskip -1.5mm}

\definecolor{red}{RGB}{255, 117, 115}
\definecolor{green}{RGB}{171, 255, 175}